\documentclass[conference]{IEEEtran}
\IEEEoverridecommandlockouts
\usepackage{cite}
\usepackage{amsmath,amssymb,amsfonts}
\usepackage{algorithmic}
\usepackage{graphicx}
\usepackage{textcomp}
\usepackage{xcolor}
\usepackage{amssymb}
\usepackage{graphicx}
\usepackage{float}
\usepackage{subfig}
\usepackage{overpic}
\usepackage{hyperref}
\hypersetup{
    colorlinks=true,
    linkcolor=blue,
    filecolor=magenta,      
    urlcolor=cyan,
    pdftitle={Overleaf Example},
    pdfpagemode=FullScreen,
    }

\def\BibTeX{{\rm B\kern-.05em{\sc i\kern-.025em b}\kern-.08em
    T\kern-.1667em\lower.7ex\hbox{E}\kern-.125emX}}
    \makeatletter
\newcommand{\linebreakand}{%
  \end{@IEEEauthorhalign}
  \hfill\mbox{}\par
  \mbox{}\hfill\begin{@IEEEauthorhalign}
}
\makeatother

\begin{document}

\title{A Novel Approach to Grasping Control of Soft Robotic Grippers based on Digital Twin\\
}

\author{\IEEEauthorblockN{1\textsuperscript{st} Tianyi Xiang}
\IEEEauthorblockA{\textit{ School of Advanced Technology} \\
\textit{Xi'an Jiaotong-Liverpool University}\\
Suzhou, China \\
Tianyi.Xiang20@student.xjtlu.edu.cn}
\and
\IEEEauthorblockN{2\textsuperscript{nd} Borui Li}
\IEEEauthorblockA{\textit{ School of Advanced Technology} \\
\textit{Xi'an Jiaotong-Liverpool University}\\
Suzhou, China \\
Borui.li20@student.xjtlu.edu.cn}
\and
\IEEEauthorblockN{3\textsuperscript{rd} Quan Zhang*(Corresponding Author)}
\IEEEauthorblockA{\textit{ School of Advanced Technology} \\
\textit{Xi'an Jiaotong-Liverpool University}\\
Suzhou, China\\
Quan.Zhang@xjtlu.edu.cn}
\linebreakand
\IEEEauthorblockN{4\textsuperscript{th} March Leach}
\IEEEauthorblockA{\textit{ School of Advanced Technology} \\
\textit{Xi'an Jiaotong-Liverpool University}\\
Suzhou, China \\
Mark.Leach@xjtlu.edu.cn}
\and
\IEEEauthorblockN{5\textsuperscript{th} Enggee Lim}
\IEEEauthorblockA{\textit{ School of Advanced Technology} \\
\textit{Xi'an Jiaotong-Liverpool University}\\
Suzhou, China \\
enggee.lim@xjtlu.edu.cn}
}

\maketitle

\begin{abstract}

This paper has proposed a Digital Twin (DT) framework for real-time motion and pose control of soft robotic grippers. The developed DT is based on an industrial robot workstation, integrated with our newly proposed approach for soft gripper control, primarily based on computer vision, for setting the driving pressure for desired gripper status in real-time. Knowing the gripper motion, the gripper parameters (e.g. curvatures and bending angles, etc.) are simulated by kinematics modelling in Unity 3D, which is based on four-piecewise constant curvature kinematics. The mapping in between the driving pressure and gripper parameters is achieved by implementing OpenCV based image processing algorithms and data fitting. Results show that our DT-based approach can achieve satisfactory performance in real-time control of soft gripper manipulation, which can satisfy a wide range of industrial applications. 

\end{abstract}

\begin{IEEEkeywords}
Digital Twin, Pneumatic flexible Actuator, OpenCV, Unity3D
\end{IEEEkeywords}

\section{Introduction}

\begin{figure*}
    \centering
    \includegraphics[width = \linewidth]{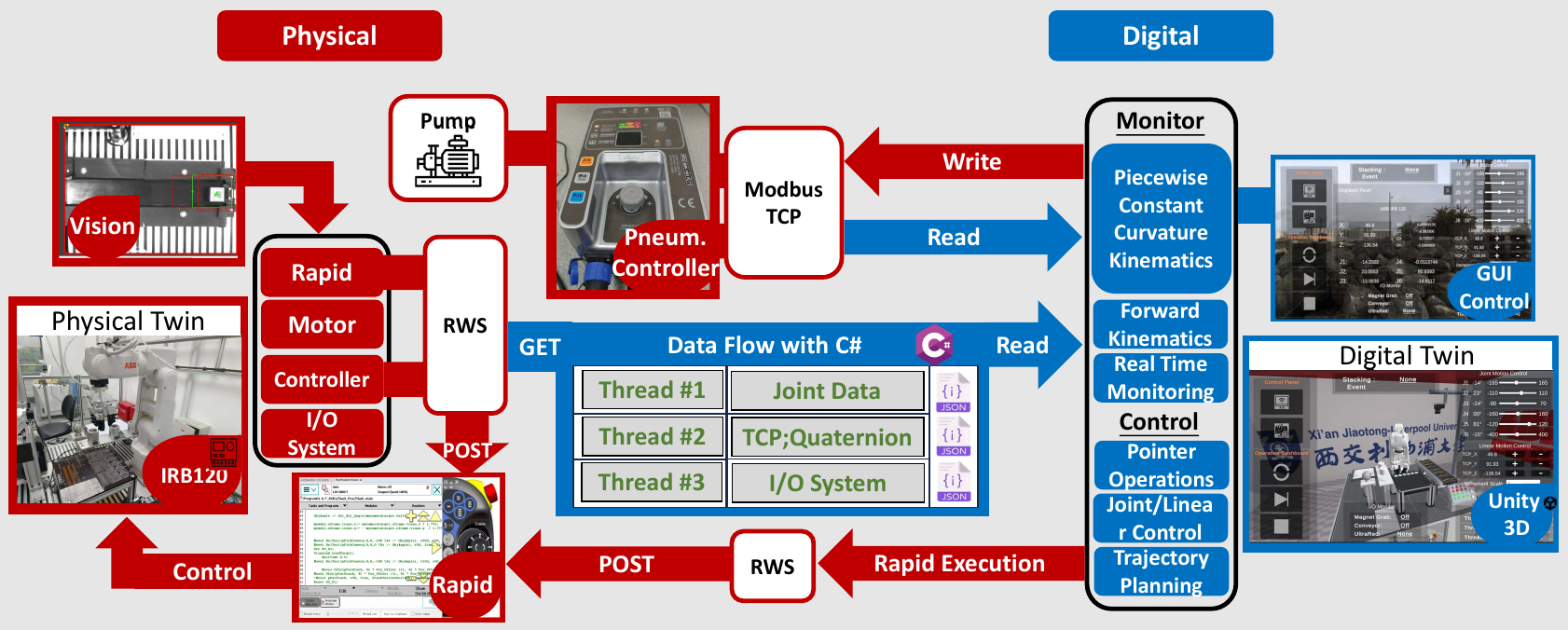}
    \caption{The Framework of the Digital Twin System for the Pneumatic Soft Solution Robot Workstation. RWS means the Robot Web Service, with its detail presented in \cite{robot}.}
    \label{Framework}
\end{figure*}

Soft robotic grippers are now experiencing a growing interest in grasping tasks, due to their considerable flexibility that allows them to grasp a variety of objects \cite{terrile2021comparison}. Such grippers are normally made of elastomeric polymers \cite{priya2022design}, with their most remarkable characteristic in that they can manipulate soft and fragile objects without damaging them 
\cite{terrile2021comparison}\cite{priya2022design}\cite{meenakshi2023fem}. In contrast to traditional rigid grippers usually designed for a specific object, flexible robotic grippers have already demonstrated their potential in various applications, e.g. food handing and packaging, fabrics handling, medical applications, etc. \cite{jin2020triboelectric}.

In real applications, soft grippers are normally required to have the capability of using a single gripper to handle various objects with different geometry and materials, rather than grasping just objects with one particular shape or material as what rigid grippers do. This requires real-time control of the gripper pose of manipulating different objects, e.g. open and closure of the gripper, the curvature and stiffness of the elastomeric fingers, etc. An approach for effective and precise  control is necessary, for real-time settings of the driving force, typically air supplying pressure to relevant pneumatic system actuating a soft gripper, of obtaining desired curvature and stiffness of gripper fingers and, subsequently, the position and orientation of the soft gripper.

Existing approaches for pose control of soft grippers mostly consider integrating the trioelectric nanogenerator (TENG) based sensors to the grippers which can detect the contact separation, continuous motion (sliding), curvature, stuffiness in a real-time manner, with the aid of related AI based algorithms for interpretation of data collected from the sensors \cite{jin2020triboelectric}. However, we believe that such approaches render the gripper design to be very complicated and, hence, less reliable, nor practical. This is because sensing of various quantities relevant to a soft gripper requires a variety of sensors, with loads of circuitry integrated in limit space with a soft gripper, and such complex system can fail easily under severe operating conditions in industrial applications. Hence, very few evidence show that such grippers with TENG-based sensors has been commercialized. Currently, commercialized soft grippers are mostly designed with elastomeric fingers driven by pneumatic power, without integration of any tactile sensors in the gripper. For real-time control of these grippers, applying computer vision appears to be an ideal solution to set the driving pressure for desired gripper pose and motion in real-time, which forms the subject matter of the present study.

The present study, therefore, aims to develop a novel approach for real-time control of soft grippers primarily based on computer vision and gripper kinematics. The real-time pose or curvature variations of the gripper fingers are captured by a 3D vision camera, and then calculated by implementing OpenCV based image processing algorithms. The real-time shift of motion and bending curvature parameters to gripper fingers, i.e. the gripper Arc parameters, can be obtained by kinematics modelling based on four-piecewise constant curvature kinematics mode \cite{Constantkinematics}. With this gripper pose parameters, the associated real-time driving pressure are then defined through data fitting.

Nowadays, digital twin (DT) technology has been extensively adopted in industrial robots in intelligent robotic systems in industry, for real-time monitoring and control of production systems and processes. We therefore have implemented our approach of soft gripper control to the DT of an industrial robot workstation developed in our earlier work, which is now equipped with a commercial soft robotic gripper. The feasibility of our approach in DT-based control and monitoring of soft gripper operations has been readily verified.



\section{The Digital Twin Framework}

Our developed DT framework is based on an industrial robot workstation, now equiped with a soft gripper, as shown in Fig. 1. The DT includes mainly two components: (1) Physical twin (2) Digital twin (Unity 3D). In between the two components,  dual-directional communication and data transmission have been readily established.

\subsection{The Physical Twin}

The physical platform is based on an industrial robotic manufacturing station with ABB IRB120 robot arm, which is similar to that reported in our earlier work \cite{CASErws}. In the present study, we use a soft gripper supplied by Rochu, with fixed module V5 and finger module AF \cite{RouchuMODULE}, as shown in Fig. \ref{kinematics}. (a). It has three air chambers and one gripper head. Furthermore, the PCU2-V pneumatic controller has been selected as the actuator \cite{RouchuMODULE}. It has three primary features: (1) Modbus TCP network port with Ethernet address, and high-speed data paths; (2) Remote reading and overwriting key actuator parameters and settings, e.g. positive or negative pressure and pressure feedback thresholds, etc. These features enable the implementation of soft gripper simulation in DT.

\subsection{The Digital Twin}

\begin{figure*}
    \centering
    \includegraphics[width = 14cm]{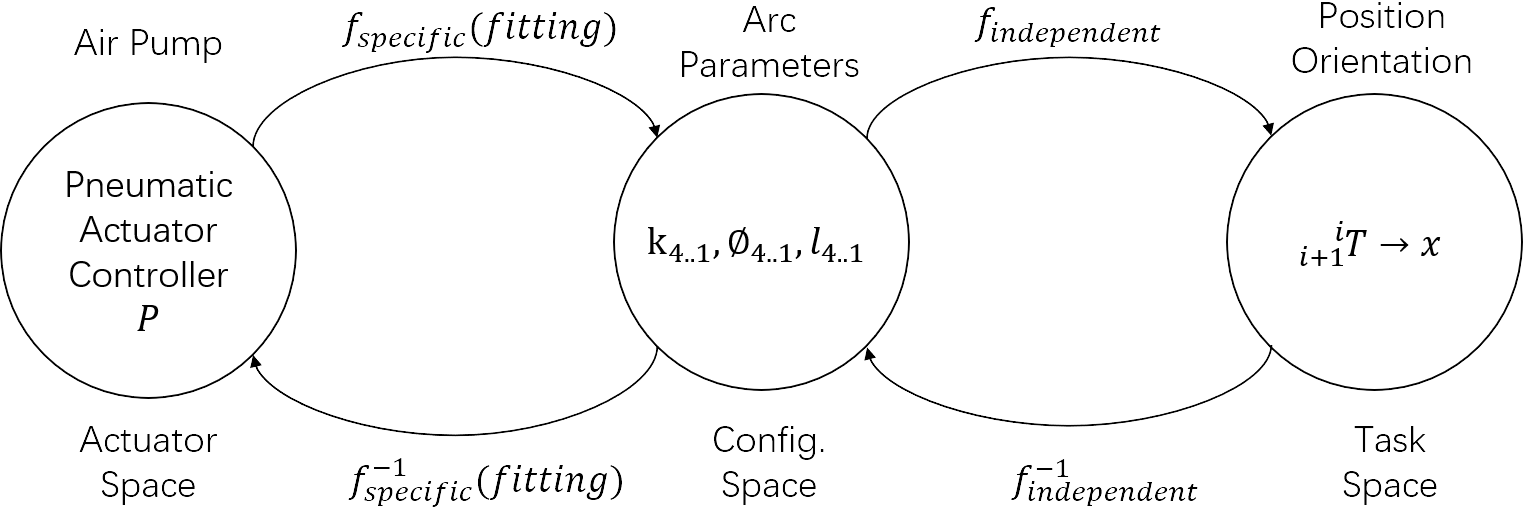}
    \caption{The triple spaces mapping among Pneumatic Actuator Controller, Arc Parameters (Configuration Space), and Position orientation (Task Space). Robot Specific Mapping: Transform pressure $P$ to configuration space variables $\mathrm{k}_{4.1}, \emptyset_{4 \ldots 1}, l_{4..1}$. Next, robot-independent mapping - Configuration Space to Task Space by kinematics}
    \label{mapping}
\end{figure*}

The data flow is mainly implemented by Robot Web Service (RWS) \cite{robot}. These streams provide key parameters for the real-time simulation with high refreshing rate. Details of dual-communication between IRB120 robot and Unity 3D can be found in our earlier paper \cite{CASErws}. The present study has modified the DT framework, adding a new pneumatic controller. The data flow read from the pneumatic controller is implemented with Modbus TCP network to ensure efficiency. The data includes: (1) positive pressure trigger state; (2) negative pressure trigger state; (3) set positive target pressure; (4) set negative target pressure (5)real-time true pressure. With the data, DT can not only simulate our developed kinematics for robot with flexible grippers but also event monitoring in real-time. Further, the DT GUI can send commands to the pneumatic controller, including: (1) overwrite positive pressure trigger state; (2) overwrite negative pressure trigger state; (3) overwrite set positive target pressure; (4) overwrite set negative target pressure. All such data can provide comprehensive control to the physical robot and pneumatic controller based on our developed user-friendly GUI.

\section{Real Time Control Approach to Flexible Actuator}

\subsection{Overall Triple Space Mapping of Piecewise Constant Curvature Kinematics for Soft Robotic Pneumatic Actuator}

The shape of our soft gripper has three air chambers and one gripper head as shown in Fig. \ref{Vision} (a). It is not feasible to consider the actuator as an entire cantilever with constant curvature. Therefore, we develop a Four-Piecewise Constant Curvature Kinematics model. The soft gripper is considered to be cut to four-piecewise sections Arc as shown in Fig. \ref{kinematics} (a), (b). It has the advantage of enabling kinematics to be decomposed into simpler three spaces mapping, as shown in Fig. \ref{mapping}. The three spaces include: (1) Pneumatic Actuator Controller; (2) Arc Parameters (Configuration Space); (3) Position Orientation (Task Space). The first mapping $f_{\text{specific}}$ is from Pneumatic Actuator space pressure, $P$, to configuration space that contains quadruple-space curves that describe position and orientation \cite{sunjie3}, which is defined as $\mathrm{k}_{4.1}, \emptyset_{4 \ldots 1}, l_{4..1}$. However, this mapping shows the main difficulty. Because there is no clear relationship between the pressure to the Arc Parameters for our wield-shape soft gripper. (a). As a result, this paper has developed a novel visual method to find mapping $f_{\text{specific}}$ in Continuum Robots by 3D depth camera vision calibration through OpenCV and data fitting. In contrast, the second mapping $f_{\text{independent}}$ transforms arc parameters to Position Orientation. This is a forward kinematics mapping implemented both in model simulation and pure mathematical in DT environment. 

Now, we first describe our novel way of deriving Mapping $f_{\text{specific}}$ for wield shape soft gripper based on OpenCV Visual calibration, then mapping $f_{\text{independent}}$ via model and pure mathematical simulation on Unity 3D.

\subsection{The Specific Mapping via Visual-based Calibration}
\begin{figure*}
    \centering
    \includegraphics[width = \linewidth]{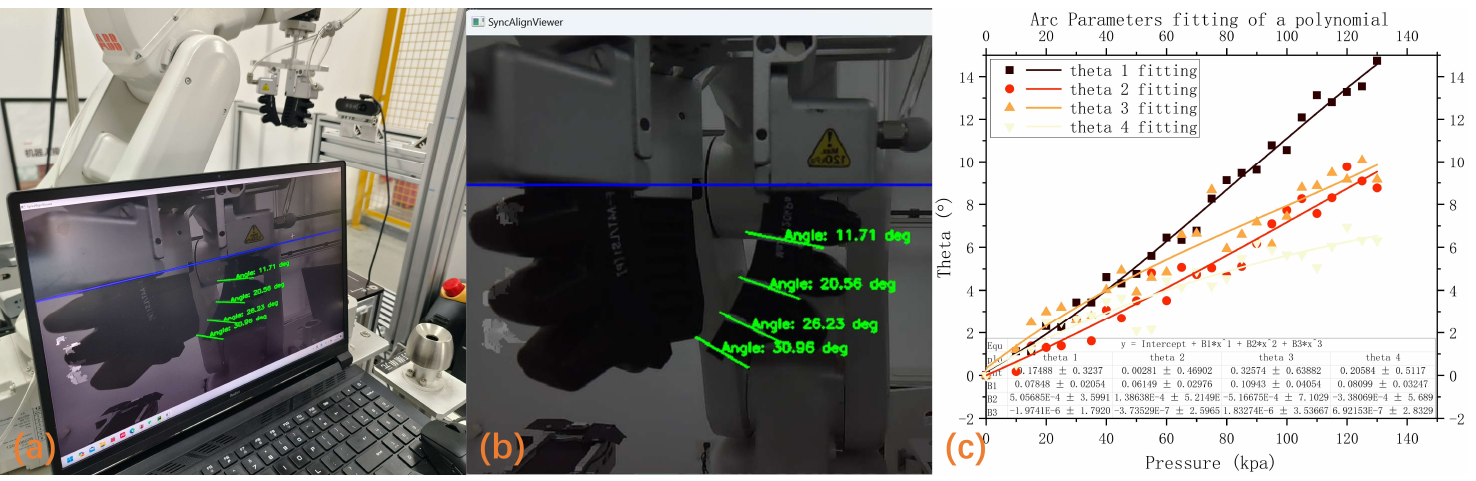}
    \caption{The 3D Vision Calibration Method to derive Specific Mapping for continuum flexible actuator: (a): The Physical Layout of Soft gripper, 3D depth Camera, and Laptop; (b) The calculated Arc Parameters $theta_{4..1}$ in calibrated world coordinates when Pressure $P = 100 Kpa$; (c) The triple polynomial fitting of Arc Parameters $theta_{4..1}$ versus Pressure $P$.}
    \label{Vision}
\end{figure*}

The specific mapping from Actuator space to Arc Parameters is computed using OpenCV Visual-based calibration. As shown in Fig. \ref{Vision}. (a), (b), we employ the Orbbeca Gemini Pro 3D depth camera in front of the soft gripper while maintaining the camera and soft gripper in the horizontal line (We can consider the camera coordinate as world coordinate). The Orbbec Gemini camera can provide depth and color streams for the calibration process \cite{orbbecGeminiORBBEC}.

To find the true dimension of the gripper in world coordinates, it is necessary to do camera calibration. Firstly, (1) ColorIntrinisic $K$; (2) ColorDistortion denoted as $k_1, k_2, k_3$ (radial) and $p_1, p_2$ (tangential) can be directly obtained from Orbbeca Software Development Kit (SDK) \cite{orbecaSDK}. Next, we call two video pipelines: (1) color stream and (2) depth stream in vision calibration to get depth info $z$ and pixel coordinate \((u, v)\) as shown in Fig. \ref{Vision}. (b). We then transform the pixel coordinates \((u, v)\) to camera coordinates \((x_c, y_c, z_c)\) for a 3D depth camera.

\begin{equation}
\begin{bmatrix}
x_c \\
y_c \\
z_c
\end{bmatrix}
= K^{-1}
\begin{bmatrix}
u_{d}  \\
v_{d}  \\
1
\end{bmatrix}
\end{equation}

where $u_{d}, v_{d}$ is the distorted pixel coordinates. We can derived it from original pixel coordinates $(u, v)$ from:

\begin{equation}
\begin{aligned}
x_{d}=  &x' \left( 1 + k_1 r^2 + k_2 r^4 + k_3 r^6 \right) \\ 
&+ 2 p_1 x' y' + p_2 (r^2 +2x'^2) \\
y_{d} = &y' \left( 1 + k_1 r^2 + k_2 r^4 + k_3 r^6 \right) \\
&+ p_1 (r^2 + 2y'^2) + 2 p_2 x' y'
\end{aligned}
\end{equation}

where the $x'$, $y'$, and $r^2$ yields:
\begin{equation}
\begin{aligned}
x' &= \frac{u_d - c_x}{f_x} \\
y' &= \frac{v_d - c_y}{f_y} \\
r^2 &= x'^2 + y'^2 \\
\end{aligned}
\end{equation}

The implementation of the distortion pixel is based on OpenCV. By incorporating depth info $z$. we can get the high accuracy mapping to the camera coordinates\((x_c, y_c, z_c)\): $x_c = x_{d} \cdot z$, $ y_c = y_{d} \cdot z $, and $z_c = z$. Next, We can develop OpenCV angle calculation method to find Arc Parameters for mapping $f_{\text{specific}}$.

\begin{figure*}
    \centering
    \includegraphics[width = 16cm]{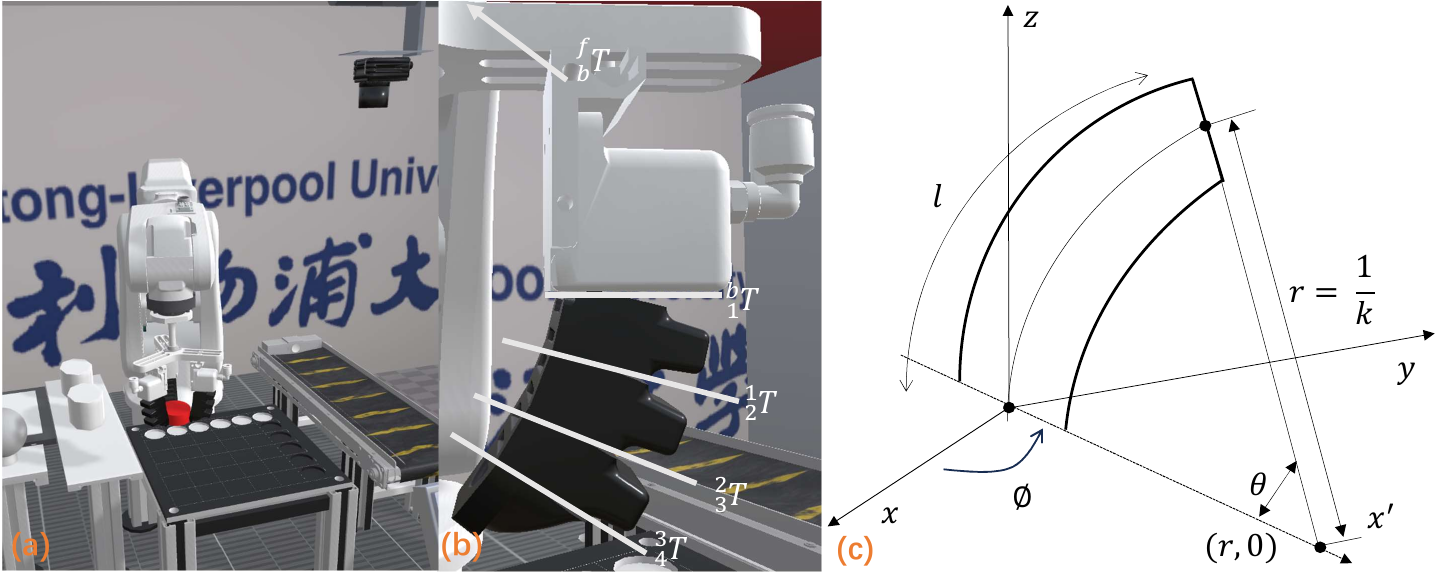}
    \caption{Four Piecewise Robot Independant Mapping Soft Gripper: (a): Flexible gripper with robot working screenshot (b): Assembled Together Four-piece Soft gripper constant curvature model with four homogeneous transformation matrix $T$ in Unity 3D; (b): Arc Configuration Space for only One piece soft gripper when angle $\phi$ rotates the arc to new x' axis.}
    \label{kinematics}
\end{figure*}

As shown in Fig. \ref{kinematics}. (b), (c), each Arc Curvature can bend with angle $\theta$ along the y-axis, and also rotates along the z-axis with $\phi$. However, the rotation $\phi$ is trivial in real industry production situations \cite{sunjienew}. Therefore, our DT simulation neglect rotation, i.e. $\phi = 0$, whilst only calibrates $\theta$ to construct the simulation by piecewise curvature $\kappa$. Nevertheless, our developed DT can not only simulate the bending with curvature $\kappa$ but also rotation along the z-axis with $\phi$, which is described in Section  \romannumeral3-C and  \romannumeral4.

Furthermore, as shown in Fig. \ref{Vision}. (b), an interactive mouse callback function was implemented to facilitate the selection of points and real-time calculation of arc parameters. By clicking on two points in the image, the system computes and displays the angle between these points relative to a predefined reference line. This interactive approach allows for efficient and intuitive determination of the gripper’s configuration. Furthermore, image processing and display the main processing loop continuously captures frames from the camera, process the depth and color data, and overlays the results on the image. As shown in Fig. \ref{Vision} (b), the fixed reference line (blue Reference line) and the lines (green lines) connecting the selected points are drawn on the image to provide a visual representation of the computed angles at the selected four cut lines between air chamber and the gripper head. These four cur lines correspond to the four sections in our developed four-section piecewise constant curvature soft gripper model and kinematics \cite{Mingjie2}.

As shown in Fig. \ref{Vision}. (c), we collect the corresponding mapping between the output $\Theta_{4..1}$ versus the input pressure $P$. The interval for each recording is $5 Kpa$. Then, We then adopt the Cubic polynomial fitting for the best fitting method \cite{mingjie1}. The detailed fitting result can be found in the table in  Fig. \ref{Vision}. (c). 

Hence, we can build the specific mapping $f_{\text{specific}}$ to configure the comprehensive space mapping in Unity 3D. Firstly, Unity 3D reads the real-time actual pressure, $p$, from the pneumatic controller to DT, i.e. the virtual twin (illustrated in Section \romannumeral2). Then, the embedded triple polynomial fitting equations $f_{\text{specific}}$ can transform the Pressure Signal to Real-time fitting Arc Parameters, $\mathrm{k}_{4.1}, \emptyset_{4 \ldots 1}, l_{4..1}$. Firstly, transform the pressure to four sections bending angles $\Theta = \begin{bmatrix}\theta_1  \theta_2  \theta_3  \theta_4\end{bmatrix}^T$.

\begin{equation}
\begin{aligned}
\kappa &= \frac{\Theta}{l} ; \Theta &= \text{intercept} + B \cdot P  \\
\end{aligned}
\end{equation}

where the pressure $P$ is $P = \begin{bmatrix}p  p^2  p^3\end{bmatrix}^T$, the interception $intercept$ and the detailed $B$ parameters can be found at Fig. \ref{Vision} (c) result table. Finally, we can transform $\Theta$ to curvature $\kappa$ with Arc length $ l= \begin{bmatrix}14, 14, 12.32, 15.39\end{bmatrix}^T$ in millimeter unit.

The specific mapping $f_{\text{specific}}$ has been successfully implemented, transforming real-time input pressure in Actuator Space to Arc Parameters $\mathrm{k}_{4.1}, \emptyset_{4 \ldots 1}$ in Configuration Space through a series of transformations and calculations based on visual data.

\subsection{The Independent Mapping Simulation in DT (Unity 3D)}

In addition to specific mapping $f_{\text{specific}}$, we need also to add the Independent Mapping $f_{\text{Independent}}$. This includes two parts: (1) Pure Mathematical Simulation for Unity3D GUI; (2) Model Kinematics Simulation.

\subsubsection{Pure Mathematical Simulation}

The simplified arc configuration space for one piece is shown as Fig. \ref{kinematics}. (c). The simplified motion shows the centered point of a circular arc with radius $r$ at $\begin{bmatrix}r,0,0\end{bmatrix}^T$ along the x-axis. The arc can rotate along the y-axis about angle $\theta$. According to the Piecewise Constant curvature kinematics \cite{Constantkinematics}, the curvature $\kappa$ indicates uniform distribution along the fixed length $l$. Note that the figure shows is the motion that rotates along the z-axis about $\phi$ angle to the new $x'$ axis. The rotation angle $\phi$ is not general at the industrial solution, so the mapping $f_{\text{specific}}$ only considers the fitting that transforms pressure $P$ to piecewise curvatures $k$ \cite{sunjie1}. However, our developed DT kinematics method can also simulate the rotation about $\phi$ angle. The transformation matrix $T$ for the one simplified piece Arc yields:

\begin{equation}
\mathrm{T}=\left[\begin{array}{cccc}
\cos \phi \cos \kappa l & -\sin \phi & \cos \phi \sin \kappa l & \frac{\cos \phi(1-\cos \kappa l)}{\kappa} \\
\sin \phi \cos \kappa l & \cos \phi & \sin \phi \sin \kappa l & \frac{\sin \phi(1-\cos \kappa l)}{\kappa} \\
-\sin \kappa l & 0 & \cos \kappa l & \frac{\sin \kappa l}{\kappa} \\
0 & 0 & 0 & 1
\end{array}\right] 
\end{equation}

Next, we assemble Four-piece sections together with four homogeneous transformation matrices $T$ shown in Fig. \ref{kinematics}. (b). The overall independent mapping $f_{\text{independent}}$ that end head of the soft gripper relative to the Robot flange center ${}_{4}^{f}T$ can be obtained by multiplying transformation matrix:

\begin{equation}
f_{\text{independent} } =  {}_{4}^{f}T = {}_{b}^{f}T {}_1^{b}T {}_2^1T{}_3^2T {}_4^3T 
\end{equation}

where Arc Length $ l= \begin{bmatrix}14, 14, 12.32, 15.39\end{bmatrix}^T$. Next, we use the real-time data flow in DT framework that (1) Tool of Center Point (TCP) of Robot flange $\begin{bmatrix} t_x, t_y, t_z \end{bmatrix}^T$; (2) Quaternion $\begin{bmatrix} q_w, q_x, q_y, q_z \end{bmatrix}^T$ to configure the robotic Arm Task Space $T_{R}$. The $T_{R}$ is the transformation matrix that the TCP center of flange relative to the robot base origin.

\begin{equation}
T_{R} = 
\begin{bmatrix}
1 - 2q_y^2 - 2q_z^2 & 2q_xq_y - 2q_zq_w & 2q_xq_z + 2q_yq_w & t_x \\
2q_xq_y + 2q_zq_w & 1 - 2q_x^2 - 2q_z^2 & 2q_yq_z - 2q_xq_w & t_y \\
2q_xq_z - 2q_yq_w & 2q_yq_z + 2q_xq_w & 1 - 2q_x^2 - 2q_y^2 & t_z \\
0 & 0 & 0 & 1
\end{bmatrix}
\end{equation}

Finally, we can derive the ultimate transformation matrix that the final end of the soft gripper relative to the origin base of the robot, $T_{F}$:

\begin{equation}
    T_{F} = T_{R} {}_{4}^{f}T
\end{equation}

\begin{figure*}
    \centering
    \includegraphics[width = 14cm]{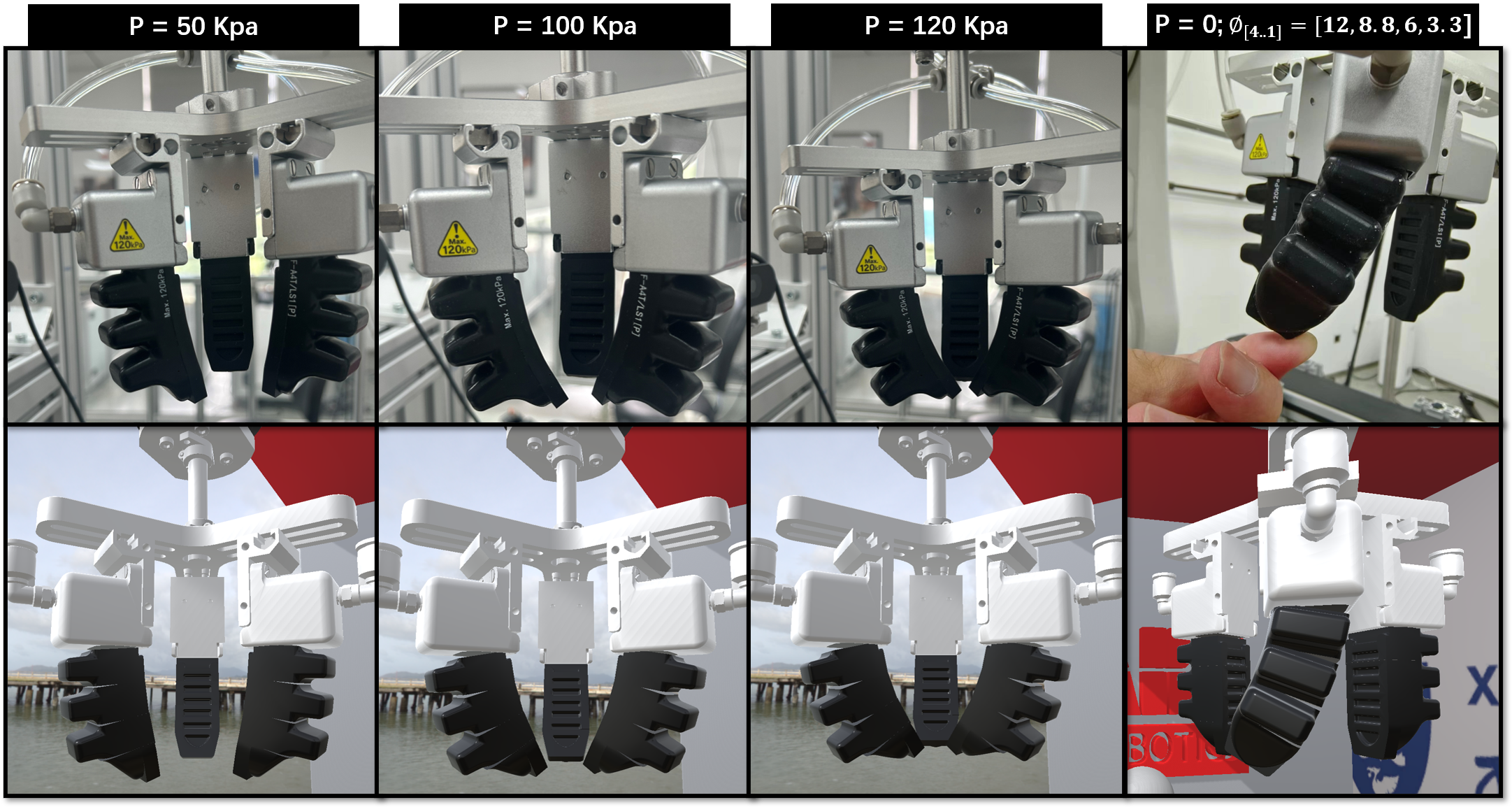}
    \caption{Bending performance mapping between reality and DT performance about curvatures $\kappa_{4..1}$ versus varying pressure; The fourth picture is human-forced simulation with $\phi = [12,8.8,6,3.3]$ that do not actually exist in real production}
    \label{bending mapping}
\end{figure*}

\subsubsection{Model Kinematics Simulation}

Real-time DT model synchronization kinematics all rely on the Transform.localEulerAngles class in Unity 3D \cite{unity3dUnityScripting}. This function relies on the Vector3 data type. It has the advantage that represent a three-dimensional rotation by performing three separate rotations around individual axes \cite{mingjie1}. For example, the Unity object rotation along the Z axis, the X axis, and the Y axis relative to its parent object's transform. Therefore, the continuum flexible gripper is cut into four pieces while set the transform at the bottom center of each section of the soft gripper. Next, we set each section's parent relationship to the up level section. Thus, we can construct the multiple-rotation that not only angle $\theta$ along the y-axis but also angle $\phi$ along the z-axis in Fig.\ref{mapping} (b) to every section of the model. 


\section{Results and Analysis}

The first three groups in Fig. \ref{bending mapping} reveal the bending performance mapping between reality and DT performance of the soft pneumatic gripper under varying pressures: $50 KPa$, $100 KPa$, and $120 KPa$). The fourth group shows that human-forced simulation with $\phi = [12,8.8,6,3.3]$, which is not general in real production. Apparently, all the bending performances exhibit high similarity. Notably, the $\phi$ rotation is trivial without simulation consideration in real Industry Production. However, our developed DT kinematics can not only simulate the bending curvature $\kappa$ but also handle the uncommon performance with $\phi$ rotation. Therefore, it proves the successful implementation of our developed four-piece constant curvature kinematics DT (Unity 3D) simulation. Next, we use Primex 41 motion capture cameras \cite{Motioncapturecamera} to capture the end head position as the truth value. Tab. \ref{bending mapping} reveals the DT mapping accuracy indicated by the soft gripper end error $E$ in Task Space Versus varying Pressure $P$ in pure mathematical simulation. $E$ defines the difference in between the simulated curvature $k$ and the true value of $k$ capture by the camera. While it performs exceptionally well $E = 0.8\%$ at lower pressure $50 kPa$, there is a little increase in error at higher pressures ($-90Kpa$ and $120kpa)$. In our DT simulation, all the bending mappings are within the expected range of the error which can satisfy most industrial applications.

\begin{table}[]
\centering
\caption{The soft gripper end error $E$ in task space versus varying pressure $P$}
\label{errortable}
\begin{tabular}{c|c}
Pressure $P$ & Task Space Error $E$ \\ \hline
$-90 Kpa$    & 2.3\%              \\
$50 Kpa $    & 0.8\%              \\
$100 Kpa $   & 1.5\%              \\
$120 Kpa $   & 3.4\%             
\end{tabular}
\end{table}

These findings have readily verified the accuracy of our approach, which is based on four-piece constant curvature kinematics including: (1) the specific mapping $f_{\text{specific}}$ with visual-based calibration by OpenCV and data fitting (Fig. \ref{Vision}); (2) the mathematical simulation and model kinematics simulation under Independent mapping $f_{\text{Independent}}$ in DT.


\section{Conclusion}

The present study has proposed an easily replicable and novel approach for real-time control of soft robotic grippers applying visual based calibration. We have also successfully adopted our approach to our previously developed DT system of a typical industrial robot workstation \cite{CASErws}, in which dual-stream communication and data transfer is implemented among (1) DT (Unity 3D), (2) robot workstation, (3) pneumatic controller, based on RWS and Modbus TCP. Then, Piecewise Constant Curvature kinematics is applied to model soft actuator into four-piece sections with independent arc parameters, while visual-data based on OpenCV with polynomial fitting is used to calibrate the triple actuator space mapping under this model. With this simple framework and kinematics, accurate real-time model visualization and mathematical mapping with GUI is achieved, with the maximum error less than $4\%$. Future work will look into more gripper parameters and errors due to hardware settings, e.g. camera errors.

\section{Acknowledgment}

The authors would like to thank the support by the XJTLU AI University Research Centre, Jiangsu Province Engineering Research Centre of Data Science and Cognitive Computation at XJTLU and SIP AI innovation platform (YZCXPT2022103).

\bibliographystyle{IEEEtran}
\bibliography{digital_twin}

\begin{thebibliography}{10}
\providecommand{\url}[1]{#1}
\csname url@samestyle\endcsname
\providecommand{\newblock}{\relax}
\providecommand{\bibinfo}[2]{#2}
\providecommand{\BIBentrySTDinterwordspacing}{\spaceskip=0pt\relax}
\providecommand{\BIBentryALTinterwordstretchfactor}{4}
\providecommand{\BIBentryALTinterwordspacing}{\spaceskip=\fontdimen2\font plus
\BIBentryALTinterwordstretchfactor\fontdimen3\font minus \fontdimen4\font\relax}
\providecommand{\BIBforeignlanguage}[2]{{%
\expandafter\ifx\csname l@#1\endcsname\relax
\typeout{** WARNING: IEEEtran.bst: No hyphenation pattern has been}%
\typeout{** loaded for the language `#1'. Using the pattern for}%
\typeout{** the default language instead.}%
\else
\language=\csname l@#1\endcsname
\fi
#2}}
\providecommand{\BIBdecl}{\relax}
\BIBdecl

\bibitem{robot}
\BIBentryALTinterwordspacing
Robot web services. [Online]. Available: \url{https://developercenter.robotstudio.com/api/rwsApi/index.html}
\BIBentrySTDinterwordspacing

\bibitem{terrile2021comparison}
S.~Terrile, M.~Arg{\"u}elles, and A.~Barrientos, ``Comparison of {{Different Technologies}} for {{Soft Robotics Grippers}},'' \emph{Sensors}, vol.~21, no.~9, p. 3253, 2021.

\bibitem{priya2022design}
I.~Priya and M.~Inzamam, ``The design and development of a soft robotic gripper,'' \emph{Materials Today: Proceedings}, vol.~68, 2022.

\bibitem{meenakshi2023fem}
S.~Meenakshi, A.~Prabhakar, and P.~Pugazhenthi, ``Fem based soft robotic gripper design for seaweed farming,'' 12 2023, pp. 1--5.

\bibitem{jin2020triboelectric}
T.~Jin, Z.~Sun, L.~Li, Q.~Zhang, M.~Zhu, Z.~Zhang, G.~Yuan, T.~Chen, Y.~Tian, X.~Hou, and C.~Lee, ``Triboelectric nanogenerator sensors for soft robotics aiming at digital twin applications,'' \emph{Nature Communications}, vol.~11, no.~1, p. 5381, 2020.

\bibitem{Constantkinematics}
\BIBentryALTinterwordspacing
I.~Robert J.~Webster and B.~A. Jones, ``Design and kinematic modeling of constant curvature continuum robots: A review,'' \emph{The International Journal of Robotics Research}, vol.~29, no.~13, pp. 1661--1683, 2010. [Online]. Available: \url{https://doi.org/10.1177/0278364910368147}
\BIBentrySTDinterwordspacing

\bibitem{CASErws}
T.~Xiang, B.~Li, X.~Pan, and Q.~Zhang, ``Development of a simple and novel digital twin framework for industrial robots in intelligent robotics manufacturing,'' in \emph{2024 IEEE 20th International Conference on Automation Science and Engineering (CASE)}, 2024.

\bibitem{RouchuMODULE}
``rochu china.com,'' \url{https://www.rochu.com/}, [Accessed 05-06-2024].

\bibitem{sunjie3}
S.~Jie, G.~Hong, M.~Rahman, and Y.~Wong, ``Feature extraction and selection in tool condition monitoring system,'' in \emph{AI 2002: Advances in Artificial Intelligence}, B.~McKay and J.~Slaney, Eds.\hskip 1em plus 0.5em minus 0.4em\relax Berlin, Heidelberg: Springer Berlin Heidelberg, 2002, pp. 487--497.

\bibitem{orbbecGeminiORBBEC}
``{G}emini2,'' \url{https://www.orbbec.com}, [Accessed 06-06-2024].

\bibitem{orbecaSDK}
``Orbbecsdk,'' \url{https://vcp.developer.orbbec.com.cn}.

\bibitem{sunjienew}
J.~Sun, K.~Yao, J.~An, L.~Jing, K.~Huang, and D.~Huang, ``Machine learning and {{3D}} bioprinting,'' \emph{International Journal of Bioprinting}, vol.~9, no.~4, pp. 717--717, 2023.

\bibitem{Mingjie2}
M.~Sun, J.~Xiao, and E.~G. Lim, ``Iterative shrinking for referring expression grounding using deep reinforcement learning,'' in \emph{2021 IEEE/CVF Conference on Computer Vision and Pattern Recognition (CVPR)}, 2021, pp. 14\,055--14\,064.

\bibitem{mingjie1}
M.~Sun, J.~Xiao, E.~G. Lim, S.~Liu, and J.~Y. Goulermas, ``Discriminative triad matching and reconstruction for weakly referring expression grounding,'' \emph{IEEE Transactions on Pattern Analysis and Machine Intelligence}, vol.~43, no.~11, pp. 4189--4195, 2021.

\bibitem{sunjie1}
S.~Jie, ``The application of nonstandard support vector machine in tool condition monitoring system,'' in \emph{Proceedings. DELTA 2004. Second IEEE International Workshop on Electronic Design, Test and Applications}, 2004, pp. 295--300.

\bibitem{unity3dUnityScripting}
``local{E}uler{A}ngles,'' \url{https://docs.unity3d.com}, [Accessed 07-06-2024].

\bibitem{Motioncapturecamera}
``{I}n {D}epth optitrack,'' \url{https://optitrack.com/cameras/primex-41/}, [Accessed 09-06-2024].

\end{thebibliography}

\end{document}